\documentclass[preprint,12pt]{elsarticle}

\usepackage{xcolor}
\usepackage{booktabs}
\usepackage{amssymb}
\usepackage{textcomp}
\usepackage{amsmath}

\usepackage{hyperref}
\usepackage{array}

\journal{Computer Vision and Image Understanding}

\begin{document}

\begin{frontmatter}

\title{Attention-guided Feature Distillation for Semantic Segmentation}

\author[1]{Amir M. Mansourian\corref{*}}  
\ead{amir.mansurian@sharif.edu}
\author[1]{Arya Jalali\corref{*}}
\ead{arya.jalali79@sharif.edu}
\author[1]{Rozhan Ahmadi}
\ead{roz.ahmadi@sharif.edu}
\author[1]{Shohreh Kasaei} 
\ead{kasaei@sharif.edu} 

\cortext[*]{Authors contributed equally.} 
\affiliation[1]{organization={Department of Computer Engineering, Sharif University of Technology},
            city={Tehran},
            country={Iran}}

\begin{abstract}
Deep learning models have achieved significant results across various computer vision tasks. However, due to the large number of parameters in these models, deploying them in real-time scenarios is a critical challenge, specifically in dense prediction tasks such as semantic segmentation. Knowledge distillation has emerged as a successful technique for addressing this problem by transferring knowledge from a cumbersome model (teacher) to a lighter model (student). In contrast to existing complex methodologies commonly employed for distilling knowledge from a teacher to a student, this paper showcases the efficacy of a simple, yet powerful method for utilizing refined feature maps to transfer attention. The proposed method has proven to be effective in distilling rich information, outperforming existing methods in semantic segmentation as a dense prediction task. The proposed Attention-guided Feature Distillation (AttnFD) method employs the Convolutional Block Attention Module (CBAM), which refines feature maps by taking into account both channel-specific and spatial information content. Simply using the Mean Squared Error (MSE) loss function between the refined feature maps of the teacher and the student, AttnFD demonstrates outstanding performance in semantic segmentation, achieving state-of-the-art results in terms of improving the mean Intersection over Union (mIoU) of the student network on the Pascal VOC 2012, Cityscapes, COCO, and CamVid datasets. Code is available at: \textnormal{\href{https://github.com/AmirMansurian/AttnFD}{https://github.com/AmirMansurian/AttnFD}}
\end{abstract}


\end{frontmatter}

\section{Introduction}
\label{sec:intro}

Semantic segmentation is a highly important and challenging task in computer vision. It has become an integral component in various applications; such as autonomous driving, video surveillance, and scene parsing. Its goal is to perform dense prediction by assigning a specific class label to each pixel in the image. Semantic segmentation has witnessed significant advancements through the use of deep neural networks, led by Fully Convolutional Network (FCN) \cite{long2015fully}. Other methods have consistently improved the segmentation accuracy by building on FCN. They achieve this by employing strategies such as designing deeper architectures to increase the capacity of FCN \cite{zhao2017pyramid}, incorporating stronger backbones \cite{huang2017densely}, and hierarchical image context processing \cite{fu2019dual, ding2024multidimensional}. Increasing the complexity is effective in improving the accuracy of semantic segmentation, yet it has become a rising concern in resource-limited environments; such as mobile and edge devices.

\begin{figure}[!ht]
	\centerline{\includegraphics[width=\textwidth]{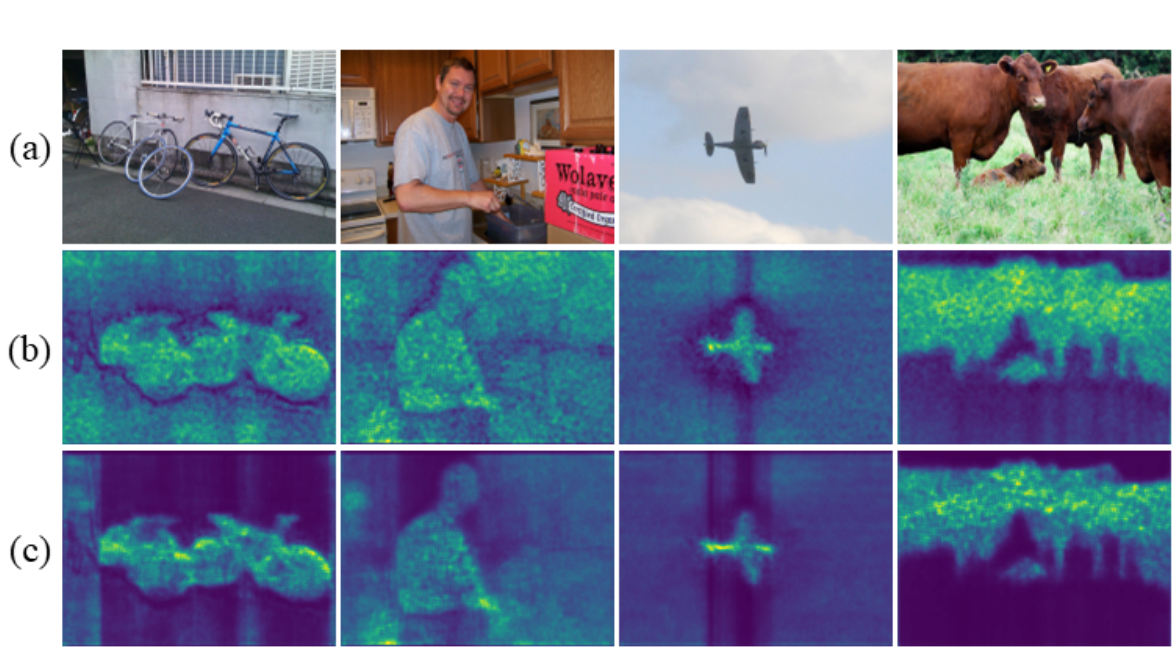}}
	\caption{Visualization of images (a), raw feature maps (b), and refined feature maps (c). Channel and spatial attention is applied to raw features, emphasizing on the important regions and making them valuable distillation source.}
	\label{fig:pull_figure}
\end{figure}

In recent years, many studies have focused on designing lightweight models with lower computational costs that are suitable for real-world applications. For example, they achieve efficiency through making the complex backbone networks lighter by reducing the number of convolutional layers or completely replacing the backbone with a simpler model. Among these methods, the  Kowledge Distillation (KD) has proven to be an effective strategy for optimizing the balance between accuracy and efficiency in deep neural networks \cite{mansourian2025comprehensive}. This technique distills useful information from a larger (teacher) network and leverages this knowledge to supervise the training of a lighter (student) network. While KD is beneficial for image classification, it faces challenges in enhancing semantic segmentation task due to its limited ability in capturing contextual correlation among pixels.

As research progresses, there has been a shift towards feature-based distillation and aligning intermediate feature maps between teacher and student networks. To achieve this, many methodologies have proposed complex loss functions to enhance knowledge distillation, since replicating feature maps with simple distance measures had limitations. While these methods are effective, recent studies \cite{liu2023simple, yang2022masked, zagoruyko2016paying} suggest that transforming student feature maps through novel modules while retaining basic loss functions can lead to simpler networks with improved performance. 

Aside from distillation, the attention mechanism in computer vision is a method that extracts a feature map and highlights its critical regions. Several attention mechanisms have been proposed recently \cite{park2018bam, woo2018cbam, qin2021fcanet, ouyang2023efficient, zhong2023lsas, ding2025attention}. Given attention's capability in refining a raw feature map, its integration into feature-based knowledge distillation holds promise for significant impact.

Unlike previous works, which either define complex losses to consider pairwise relations or rely on raw features, this study leverages the attention mechanism in CBAM\cite{woo2018cbam}. This mechanism incorporates both channel and spatial information to produce refined features, which are then distilled from the teacher to the student. Figure \ref{fig:pull_figure} illustrates the distinction between raw and refined features. As depicted in this figure, the refined feature highlights important regions of the image and reduces background noise, making it a strong candidate with significant potential for distillation. This is because it compels the student network to mimic the important regions emphasized by the teacher. A summary of the main contributions of this work includes: 
\begin{itemize}
	
        \item Investigating the role of attention mechanisms in feature-based distillation methods for semantic segmentation.

        \item Proposing a novel and simple attention-based feature distillation method using the CBAM attention module. This is the first work to use both channel and spatial attention in the context of knowledge distillation.

        \item Validating the effectiveness of the proposed method by comparing it to existing state-of-the-art methods across four widely used benchmarks with two different network architectures.
	
\end{itemize}

\section{Related Work}
A literature review of state-of-the-art studies relevant to the proposed method is presented in this section. It contains discussions on KD and attention mechanism.

\subsection{Knowledge Distillation}

Initially introduced in \cite{hinton2015distilling}, the concept of KD is to minimize the Kullback-Leibler Divergence between probability maps of the teacher and the student. Subsequently, various paradigms emerged. Fitnet \cite{romero2014fitnets} extracts and aligns feature maps from the hidden layers of a network. In \cite{zagoruyko2016paying} the student undergoes training to emulate the corresponding intermediate attention map from the teacher. 
 The RKD \cite{park2019relational} involves extracting distance and angle-based correlations between feature maps. The work in \cite{peng2019correlation} presents a framework to balance the correlation and instance compatibility between samples. The \cite{zhao2021similarity} provides the student model with a softer sample distribution through a mixture of input samples. 
On the other hand, certain studies attempt to improve the KD through various tricks, such as employing an adaptive cross-entropy loss function \cite{park2020knowledge}, reducing the duration of distillation's impact \cite{zhou2020channel, mansourian2025aicsd}, pruning features before the distillation process \cite{park2022prune} or investigating the role of projectors and temperature in KD \cite{miles2024understanding, li2023curriculum}.

\begin{figure}[ht]
	\centerline{\includegraphics[width=\textwidth]{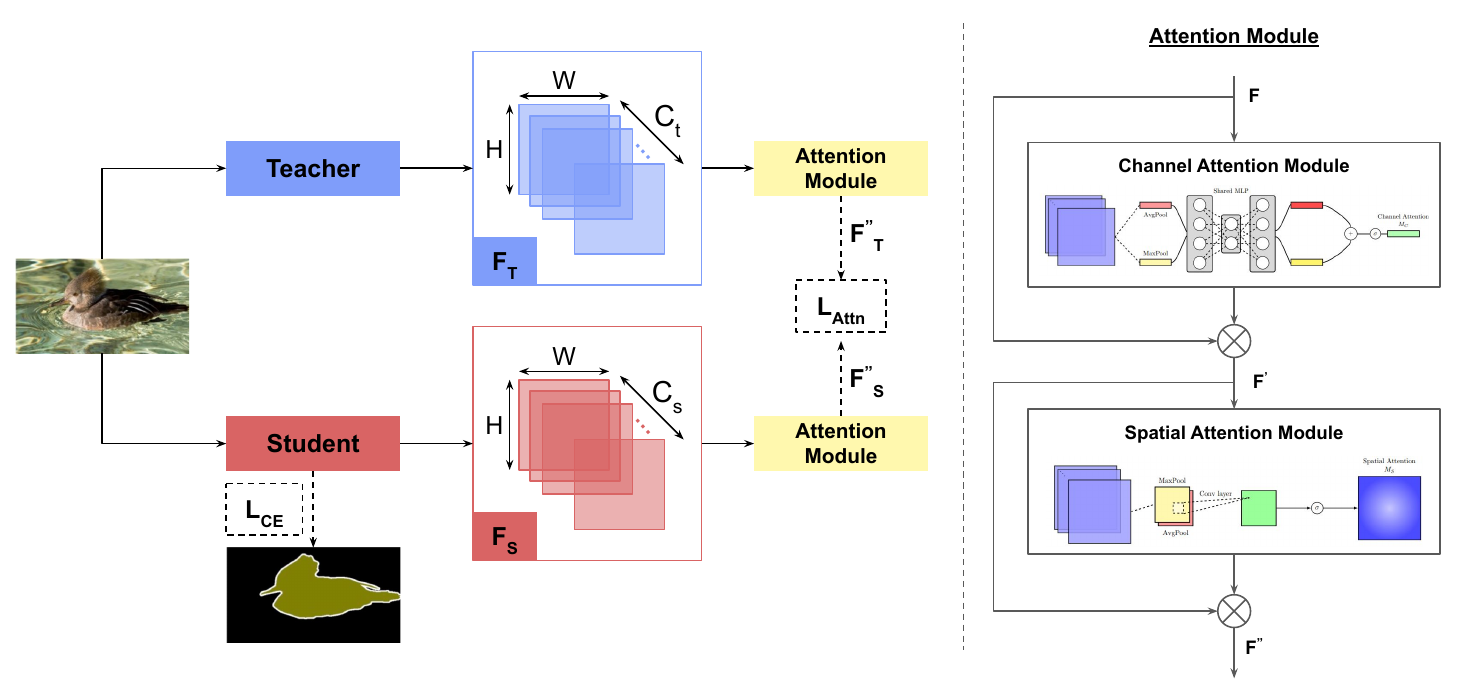}}
	\caption{Overall diagram of the proposed distillation method. The student model is trained with cross-entropy loss ($L_{ce}$) along with distillation loss between the refined feature maps of the teacher and the student ($L_{attn}$). Refined feature maps are obtained by applying channel attention, followed by spatial attention on the feature maps. The teacher's parameters are frozen during the training of the student, and any inconsistency between the sizes of the features is compensated using interpolation and convolution operations.}
	\label{fig:method}
\end{figure}

\subsubsection{Knowledge Distillation for Semantic Segmentation}

For semantic segmentation, which requires high-level information, more advanced forms of knowledge distillation have been proposed, such as adversarial training \cite{liu2019structured} and prototyping \cite{wang2020intra}. Furthermore, different similarity distillation methods have been proposed at various levels, including instance-level \cite{tung2019similarity}, class-level \cite{feng2021double, huang2022knowledge}, and channel-level \cite{park2020knowledge, liu2021exploring}. More recently, additional forms of relationships have been considered. The CIRKD \cite{yang2022cross} extracts relations across images to capture a better global knowledge about pixel dependencies. The BPKD \cite{liu2024bpkd} uses separate distillation loss functions for body and edge to improve edge differentiation.

Although these methods have proven to be effective, their novel ways of defining and distilling the knowledge usually result in complex models that require prior knowledge and careful feature extraction processes. As a result, some studies have gravitated towards designing modules to transform features and extract rich information from them. The proposed method in MGD \cite{yang2022masked} involves masking random pixels of the student's feature and training it to replicate the feature of the teacher. Some recent works have also showcased improved performance by using the raw features directly or through simple transformations. The MLP \cite{liu2023simple} achieves this by aligning features across their channel dimension, using a simple channel-wise transformation. The DiffKD \cite{huang2024knowledge} treats the student's feature maps as a noisy version of the teacher's feature maps, and trains a diffusion model to refine the student's features for more effective knowledge distillation. The LAD \cite{liu2024rethinking} demonstrates that using the MSE loss between raw features of the teacher and the student can significantly improve the performance.

\subsection{Attention Mechanism}
This mechanism allows models to extract local context information from an image more accurately \cite{vaswani2017attention}. Some methods, such as SE-NET \cite{hu2018squeeze} and SGE-NET \cite{li2019spatial}, calculate attention at the inter-channel level, and Self-attention \cite{ambartsoumian2018self} considers pair-wise similarity of the input's pixels. The BAM \cite{park2018bam} calculates channel and spatial attention, CBAM \cite{woo2018cbam} performs spatial attention after channel-wise attention, and FcaNet\cite{qin2021fcanet} proposes a channel attention mechanism in the frequency domain. Recently, EMA\cite{ouyang2023efficient} have proposed an efficient multi-scale attention by channel grouping and Lsas\cite{zhong2023lsas} have proposed a lightweight Sub-attention Strategy for alleviating attention bias problem. 

In the context of KD, several works have employed the attention mechanism. \cite{an2022efficient} utilizes an attention mechanism by incorporating the concept from \cite{fu2019dual} and employing spatial self-attention for the distillation process. \cite{yuan2024student} uses self-attention to simplify the learning process by softening the logits before distillation, \cite{karine2024channel} applies channel self-attention and position self-attention on the features for distillation, and SAKD\cite{guo2024sakd} computes attention using student features as queries and teacher features as key values, implementing sparse attention values through random deactivation. 

In this work, we employ the CBAM attention mechanism to refine raw features, as it adaptively performs the spatial and channel attention. To the best of our knowledge, this is the first instance of utilizing both channel and spatial attention for the purpose of knowledge distillation.

\section{Methodology}
This section first reviews feature distillation and then presents the details of our proposed method.

\subsection{Revisiting Feature Distillation}
In feature distillation, the student network is trained to mimic the teacher's feature maps:
\begin{align}
	\label{feature distillation}
	L_{fd} = \ell_{feat}(\Phi_s(F_s), \Phi_t(F_t))
\end{align}
\noindent where $\ell_{feat}$ is a similarity function for matching the teacher's ($F_t$) and student's ($F_s$) features, and $\Phi$ is a transformation used to align the feature size and channels if there is an inconsistency.

Recent studies have suggested using more sophisticated transformations ($\Phi_s, \Phi_t$) to not only align feature sizes, but also to better enable student network to acquire knowledge from teacher network. For instance, MLP \cite{liu2023simple} utilizes a simple channel transformation, and DiffKD \cite{huang2024knowledge} proposes a diffusion model for feature denoising. This research direction raises the question of whether the use of well-designed transformation is necessary for student network to learn more effective features from the teacher.

This work conducts empirical studies to address the aforementioned question, and finds that student model can boost its representations through an attention transformation. Leveraging this insight, as shown in Figure \ref{fig:method}, we introduce a straightforward method that integrates a Convolutional Block Attention Module (CBAM) \cite{woo2018cbam} into the student and teacher features, and aligns the refined student features with the refined teacher features using a conventional MSE distance. The following sections first describe the CBAM module and then detail our proposed Attention-guided Feature Distillation (AttnFD) approach.

\subsection{Convolutional Block Attention Module}

\begin{figure}[h]
	\centerline{\includegraphics[width=\textwidth]{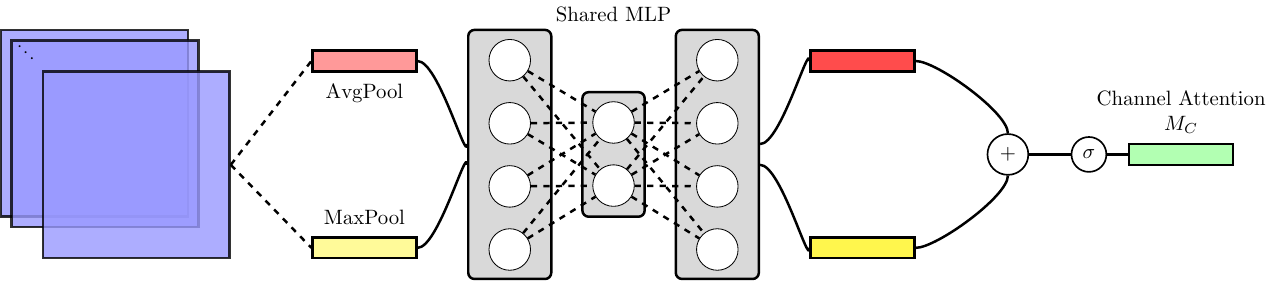}}
	\caption{Channel Attention Module. It applies average-pooling and max-pooling operators along the channel dimension. Resulting outputs are then passed through a shared Multi-Layer Perceptron and fed into a sigmoid activation function to generate the channel attention map $M_C$.}
	\label{fig:cam}
\end{figure}

\begin{figure}[h]
	\centerline{\includegraphics[width=\textwidth]{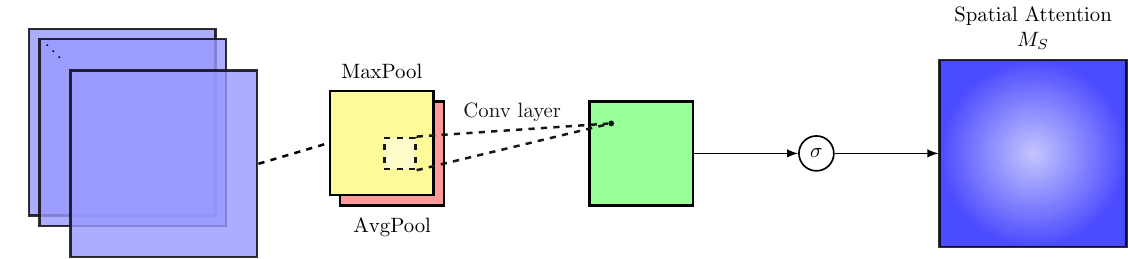}}
	\caption{Spatial Attention Module. Max-pooling and average-pooling operators are utilized to generate feature descriptors. These descriptors are subsequently fed into a convolution layer and a sigmoid activation function, resulting in spatial attention map $M_S$.}
	\label{fig:sam}
\end{figure}

Let $F\in \mathbb{R}^{c\times w\times h}$ be an intermediate feature map obtained from the student or the teacher network with spatial and channel dimensions of $h \times w$ and $c$, respectively. Two attention modules aggregate channel ($M_C(F) \in \mathbb{R}^{c \times 1 \times 1}$) and spatial descriptors ($M_S(F) \in \mathbb{R}^{c \times h \times w}$) of an intermediate feature. These spatial and channel attention descriptors are then multiplied by the original feature map $F$ to create new rich context feature maps $F^{'}$ and $F^{''}$, introducing inter-channel and inter-spatial information into the original feature map. The overall formulation of feature refinement is given by
\begin{align}
	\label{Feature enrichment}
	F^{'} &= M_{C}(F) \otimes F \\
	F^{''} &= M_{S}(F^{'}) \otimes F^{'} 
\end{align}
\noindent where $\otimes$ is the element-wise multiplication. During multiplication, spatial attention maps are broadcasted along the channels, and channel attention maps are broadcasted along the spatial dimensions. The methodology of the channel and spatial attention modules can be seen in Figure \ref{fig:cam} and Figure \ref{fig:sam}, respectively.

\subsubsection{Channel Attention Module}
The Channel Attention Module (CAM) aggregates spatial information by applying the max-pooling and average-pooling operators to an intermediate feature map $F$. These operations produce context descriptors, which are then processed by a multi-layer perceptron to generate a channel attention map $M_C(F)$. This map highlights the meaningful regions of the image while obscuring regions that are irrelevant to the segmentation task, such as the background.
The channel-aware descriptors for an intermediate feature map $F$ are defined by
\begin{align} \label{Channel_Attention}
	M_C(F) &= \sigma(W_1(W_0(F^C_{\text{avg}}))) + W_1(W_0(F^C_{\text{max}}))
\end{align}
\noindent where $F^C_{\text{avg}} \in \mathbb{R}^{c \times 1 \times 1}$ and $F^C_{\text{max}} \in \mathbb{R}^{c \times 1 \times 1}$ are the feature maps generated by applying the average-pooling and max-pooling operators to the intermediate feature map $F$. The $W_1$ and $W_0$ are the weights of the shared MLP between the two pooled feature maps. The $W_0$ is followed by a ReLU activation function, and $\sigma$ denotes the sigmoid function.

\subsubsection{Spatial Attention Module}
The Spatial Attention Module (SAM), shown in Figure \ref{fig:sam}, operates fairly similar to CAM. Spatial information of an intermediate feature map $F$ are aggregated by using the max and average pooling operators to generate two different spatial context descriptors $F^{S}_{\text{avg}} \in \mathbb{R}^{1 \times h \times w}$ and $F^{S}_{\text{max}} \in \mathbb{R}^{1 \times h \times w}$. Then, the spatial context descriptors for $F$ are calculated as
\begin{align}
	M_{S}(F) &= \sigma(A^{7 \times 7}([F^{S}_{\text{avg}}; F^{S}_{\text{max}}])) 
\end{align}
\noindent where $A^{7 \times 7}$ represents a $7 \times 7$ convolution kernel.

\subsection{Attention-guided Feature Distillation} Using the newly acquired features, the proposed loss is calculated as
\begin{align}
	L_{Attn} = \frac{1}{N} \sum_{i = 1}^{N} \left\Vert \frac{F^{''}_{\mathcal{S}j}}{\parallel F^{''}_{\mathcal{S}j} \parallel} - \frac{F^{''}_{\mathcal{T}j}}{\parallel F^{''}_{\mathcal{T}j} \parallel} \right\Vert
\end{align}
\noindent where $F^{''}_{\mathcal{S}j}$ and $F^{''}_{\mathcal{T}j}$ represent the $j$'th intermediate context-rich feature map for the student and teacher network, respectively. Each feature map is normalized along its channels before calculating the difference matrix.

The overall loss function is then a weighted sum of $L_{CE}$ and $L_{Attn}$, given by
\begin{equation} \label{overal_loss} 
	\centering
	L_{AttnFD} = L_{CE} + \alpha L_{Attn}
\end{equation}
\noindent where $\alpha$ is a weight coefficient which is fine-tuned as described in next section. In this context, the well-known cross-entropy loss function, $L_{CE}$, is employed as the segmentation loss between the predictions of the student network and the ground-truth labels.


\section{Experimental Results}
\subsection{Datasets and Evaluation Metrics}

(1) \textbf{Cityscapes} \cite{cordts2016cityscapes} dataset, designed for understanding urban scenes, includes 2975/500/1525 images for train/val/test, covering 19 classes.
(2) \textbf{Pascal VOC} \cite{everingham2010pascal} dataset comprises 1464/1449/1456 images for train/val/test, with 21 classes.
(3) \textbf{COCO} \cite{everingham2010pascal} dataset is a large-scale dataset, containing 118K/5K/41K labeled images for the train/val/test sets, across 91 classes.
(4) \textbf{CamVid} \cite{brostow2009semantic} is a road scene understanding dataset with a total of 701 labeled video frames, including 367/101/233 images for train/val/test sets, in 11 classes.


\textbf{Evaluation Metrics} As per the standard, segmentation performance is evaluated using the mIoU and pixel accuracy, averaged over three runs for fair comparison. The model size is indicated by the reported number of network parameters.

\begin{table}[t]
	\caption{\label{tab: pascal-results}Quantitative results on Pascal VOC Validation set.}
	\centering
    \resizebox{\linewidth}{!}{%
    \begin{tabular}{>{\raggedright\arraybackslash}m{\dimexpr 0.5\linewidth}|%
                    >{\centering\arraybackslash}m{\dimexpr 0.5\linewidth}%
                    >{\centering\arraybackslash}m{\dimexpr 0.2\linewidth}%
                    }
		\toprule

		Method &  mIoU(\%) & Params(M) \\
		\midrule 
		T: DeepLabV3-ResNet101  & 77.85  & 59.3 \\
		
		\midrule
		
		S: DeepLabV3-ResNet18 & 67.50 &  \\
		S + KD  & 69.13 $\pm$  0.11  &  \\
		S + ICKD  & 69.13 $\pm$ 0.17 &  \\
		S + DistKD  & 69.84 $\pm$ 0.11  & 16.6 \\
		S + CIRKD  & 71.02 $\pm$ 0.11 &  \\
            S + AICSD  & 70.58 $\pm$ 0.14 &  \\
		S + LAD  & 71.42 $\pm$ 0.09 &  \\
		S + AttnFD (ours)   & \textbf{73.09 $\pm$ 0.06} &  \\
		
		\midrule
		
		S: DeepLabV3-MBV2 & 63.92 &  \\
		S + KD  & 66.39 $\pm$  0.21  &  \\
		S + ICKD  & 67.01 $\pm$ 0.10 &  \\
		S + DistKD  & 67.62 $\pm$ 0.22  & 5.9 \\
		S + CIRKD  & 69.02 $\pm$ 0.16 &  \\
            S + AICSD  & 68.56 $\pm$ 0.26 &  \\
		S + LAD  & 68.63 $\pm$ 0.07 &  \\
		S + AttnFD (ours)  & \textbf{70.38 $\pm$ 0.16} &  \\
		
		\midrule
		
		S: PSPNet-ResNet18 & 67.40 &  \\
		S + KD  & 68.18 $\pm$  0.08  &  \\
            S + ICKD  & 68.21 $\pm$ 0.11 &  \\
		S + DistKD  & 68.93 $\pm$ 0.19  & 12.6 \\
		S + CIRKD  & 69.53 $\pm$ 0.11 &  \\
            S + AICSD  & 68.29 $\pm$ 0.18 &  \\
		S + LAD  & 69.71 $\pm$ 0.10 & \\
		S + AttnFD (ours)   & \textbf{70.95 $\pm$ 0.06} &  \\
		
		\bottomrule 
	\end{tabular}}
\end{table}

\subsection{Implementation Details}\label{subs:Implementation Details}
\subsubsection{Network Architecture}
To ensure impartial assessment, the experiments employ identical teacher and student networks as described in \cite{liu2021exploring}. The teacher network consistently applied across all experiments is Deeplab V3+ with ResNet-101 serving as the backbone. Various backbones, such as ResNet-18 and MobileNetv2, are utilized for the student network within the Deeplab V3+ and PSPNet segmentation networks.

\begin{figure}[!ht]
	\centerline{\includegraphics[width=\textwidth]{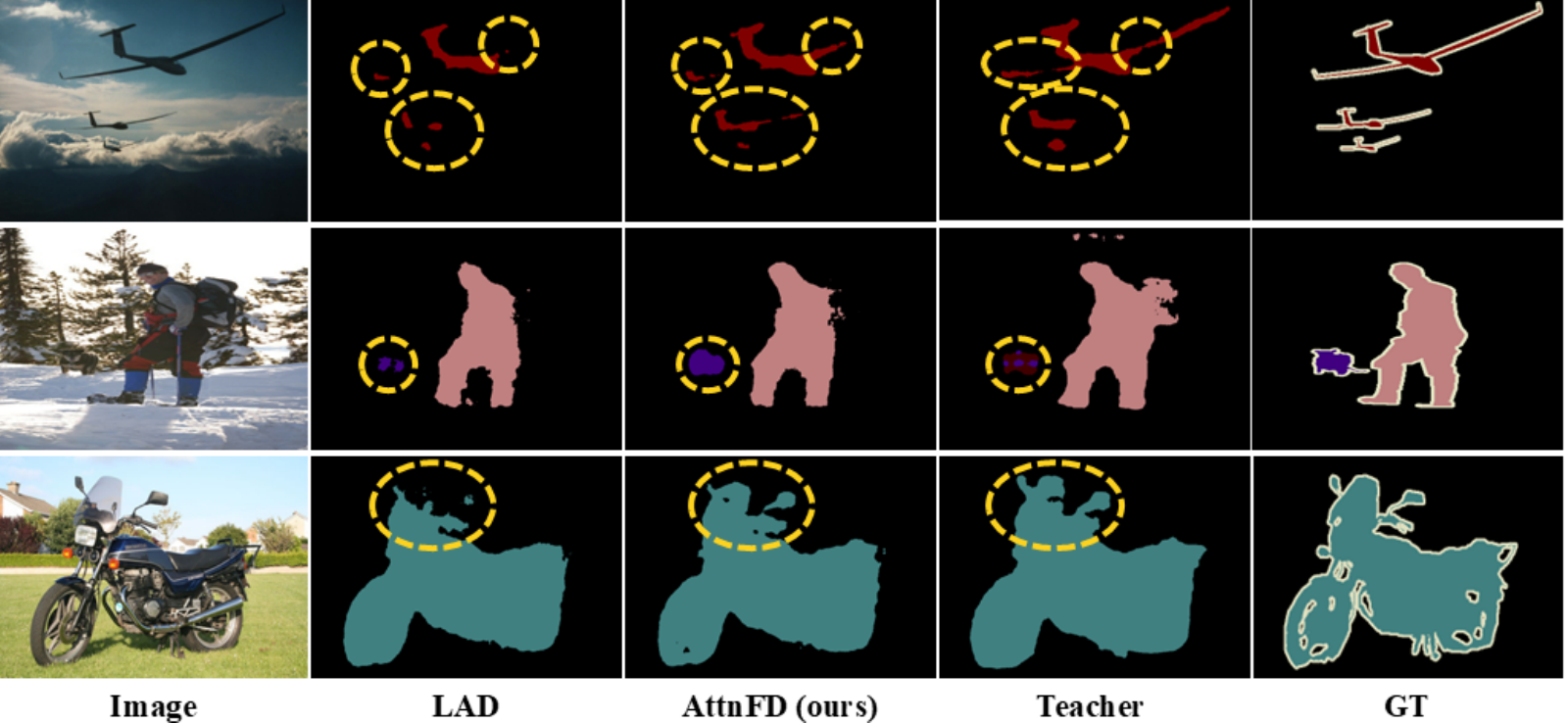}}
	\caption{Some qualitative comparisons on the Pascal VOC validation split.}
	\label{fig:pascal_vis}
\end{figure}

\subsubsection{Training Details}
All the datasets utilize similar configurations for training the student networks. For the Pascal and CamVid datasets, a batch size of 6 and a total of 120 epochs are employed, whereas for the Cityscapes dataset, a batch size of 4 and a total of 50 epochs, and for the COCO dataset, a batch size of 6 and a total of 10 epochs are utilized. The Stochastic Gradient Descent (SGD) optimizer is employed with an initial learning rate set to 0.007 for Pascal, 0.01 for Cityscapes, 0.1 for COCO, and 0.02 for CamVid. The learning rate adjustment is performed based on the cosine annealing scheduler. Before the training phase, each image undergoes preprocessing, including random scaling between 0.5 to 2 times its original size, horizontal random flipping, and random cropping to dimensions of $513\times513$ pixels for Pascal and COCO,  $512\times1024$ for Cityscapes, and $360\times360$ for CamVid. The backbones of both the teacher and student networks utilize pre-trained weights from the ImageNet dataset, while the segmentation parts are initialized randomly. The only hyperparameter of the method described in equation \ref{overal_loss} was fine-tuned by experimenting with values \{0.1, 1, 10, 100\} for finding proper value, and then trying with some different numbers around it for finding the best value. It was established as $\alpha=2$ for the Pascal, COCO, and CamVid datasets, and $\alpha=15$ for the Cityscapes dataset. During inference, performance is assessed on original inputs at a single scale. Instead of utilizing raw feature maps, Pre-ReLU feature maps are employed, as suggested in \cite{heo2019comprehensive}, to retain negative values for applying attention module. The implementation is in PyTorch framework, with all networks trained on a single NVIDIA GeForce RTX 3090 GPU.

It is important to highlight that any inconsistency in the size and number of channels of features between the teacher and the student is addressed through the interpolation and inclusion of a convolutional layer respectively. Additionally, the parameters of the teacher and its attention module are trained and then frozen during the training of the student.

\begin{table}[ht]
	\caption{\label{tab: cityscapes-results} Quantitative results on Cityscapes Validation set.}
	\centering
    \resizebox{\linewidth}{!}{%
    \begin{tabular}{>{\raggedright\arraybackslash}m{\dimexpr 0.45\linewidth}|%
                    >{\centering\arraybackslash}m{\dimexpr 0.45\linewidth}%
                    >{\centering\arraybackslash}m{\dimexpr 0.45\linewidth}%
                    }
		\toprule
		Method &  mIoU(\%) & Accuracy(\%) \\
		\midrule 
		T: DeepLabV3-ResNet101  & 77.66  & 84.05 \\
		
		\midrule
		
		S: DeepLabV3-ResNet18 & 64.09 & 74.80 \\
		S + KD  & 65.21 \textcolor{black}{(+1.12)} & 76.32 \textcolor{black}{(+1.74)} \\
		S + ICKD  & 66.98 \textcolor{black}{(+2.89)} & 77.48 \textcolor{black}{(+2.90)} \\
		S + CIRKD  & 70.49 \textcolor{black}{(+6.40)} & 79.99 \textcolor{black}{(+5.19)}\\
		S + DistKD  & 71.81 \textcolor{black}{(+7.72)} & 80.73 \textcolor{black}{(+5.93)} \\
            S + AICSD  & 70.96 \textcolor{black}{(+6.85)} & 80.24 \textcolor{black}{(+5.42)} \\
		S + LAD  & 71.37 \textcolor{black}{(+7.28)} & 80.93 \textcolor{black}{(+6.13)} \\
		S + AttnFD (ours)   &  \textbf{73.04 (+8.95)} &  \textbf{83.01 (+8.21)} \\
		
		\midrule
		
		S: DeepLabV3-MBV2 & 63.05 & 73.38 \\
		S + KD  & 64.03 \textcolor{black}{(+0.98)} & 75.34 \textcolor{black}{(+1.96)} \\
		S + ICKD  & 65.55 \textcolor{black}{(+2.50)} & 76.48 \textcolor{black}{(+3.10)} \\
		S + CIRKD  & 69.34 \textcolor{black}{(+6.39)} & 78.66 \textcolor{black}{(+5.28)} \\
		S + DistKD  & 69.53 \textcolor{black}{(+6.48)} & 79.10 \textcolor{black}{(+5.72)} \\
            S + AICSD  & 69.51 \textcolor{black}{(+6.56)} & 78.83 \textcolor{black}{(+5.45)} \\
		S + LAD  & 69.84 \textcolor{black}{(+6.79)} & 80.49 \textcolor{black}{(+7.11)} \\
		S + AttnFD (ours)   &  \textbf{70.80 (+7.75)} &  \textbf{81.59(+8.15)} \\
		
		\midrule
		
		S: PSPNet-ResNet18 & 65.72 & 73.77 \\
		S + KD  & 66.89 \textcolor{black}{(+1.17)} & 74.82 \textcolor{black}{(+1.05)} \\
            S + ICKD  & 66.98 \textcolor{black}{(+1.26)} & 74.96 \textcolor{black}{(+1.19)} \\
		S + CIRKD  & 67.51 \textcolor{black}{(+1.79)} & 75.25 \textcolor{black}{(+1.48)} \\
		S + DistKD  & 68.13 \textcolor{black}{(+2.41)} & 76.25 \textcolor{black}{(+2.48)} \\
            S + AICSD  & 67.92 \textcolor{black}{(+2.20)} & 76.01 \textcolor{black}{(+2.24)} \\
		S + LAD  & 67.71 \textcolor{black}{(+1.99)} & 75.63 \textcolor{black}{(+1.86)} \\
		S + AttnFD (ours)   &  \textbf{68.86 (+3.14)} &  \textbf{76.47 (+2.70)} \\
		
		\bottomrule 
	\end{tabular}}
\end{table}

\begin{figure}[!ht]
	\centerline{\includegraphics[width=\textwidth]{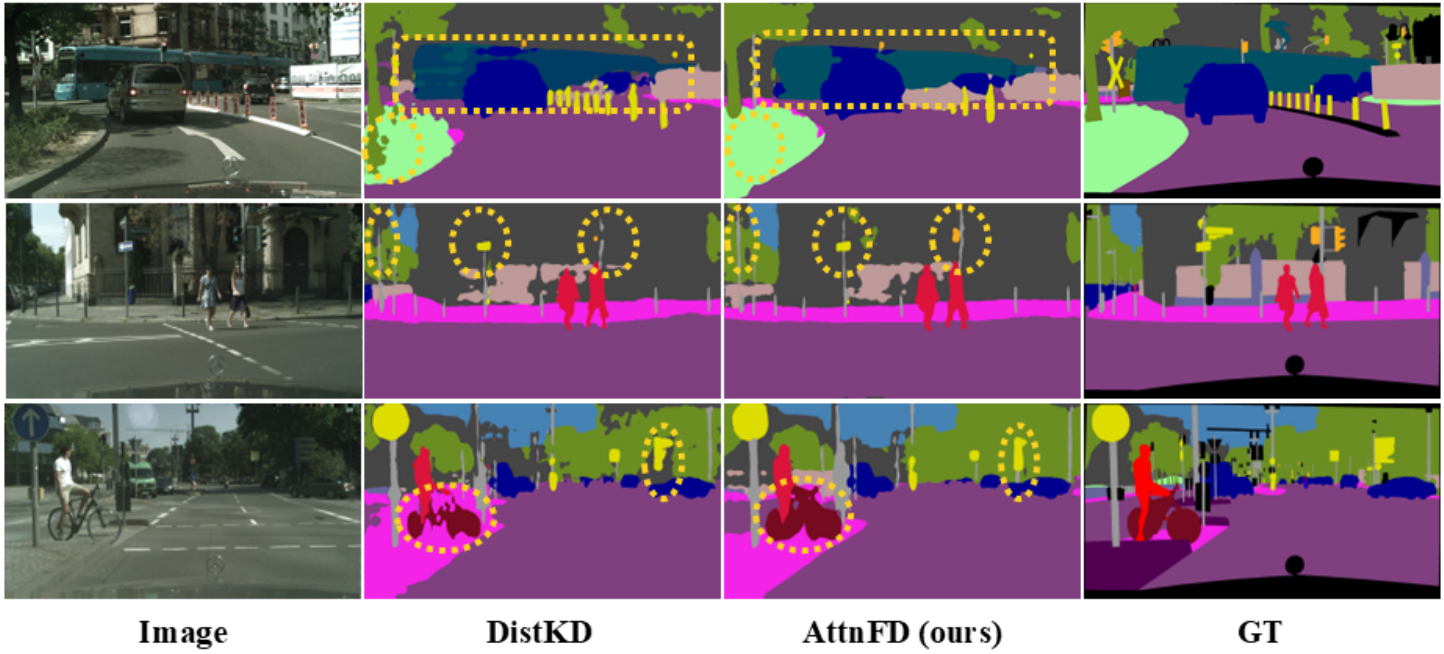}}
	\caption{Enhancements in visual quality on Cityscapes validation set.}
	\label{fig:city_vis}
\end{figure}

\subsection{Comparison with state-of-the-art methods}
Extensive experiments were conducted to assess the performance of the proposed AttnFD. It was compared against several existing distillation methods; namely, KD \cite{hinton2015distilling}, ICKD \cite{liu2021exploring}, CIRKD \cite{yang2022cross}, DistKD \cite{huang2022knowledge}, AICSD \cite{mansourian2025aicsd}, and LAD \cite{liu2024rethinking}. Each of the aforementioned methods underwent testing across all backbone, encoder, and decoder features, as well as final output maps, to determine the optimal results.

\subsubsection{Pascal VOC.}
Initially, the outcomes of the proposed method are compared with those of the aforementioned methods on the Pascal VOC dataset. As illustrated in Table \ref{tab: pascal-results}, AttnFD demonstrates notable performance enhancement for the model without distillation, achieving a 5.59\% increase when employing ResNet-18 as the student backbone, and a 6.46\% improvement with MobileNet as the student backbone. Moreover, our method surpasses the top-performing existing methods by a significant margin. Specifically, it outperforms LAD by 1.67\% with ResNet-18, and CIRKD by 1.36\% with MobileNet as the student backbone, positioning it as the second-best method. Furthermore, AttnFD outperforms LAD by 1.24\% in performance when a student model with a PSPNet architecture is employed, demonstrating that the proposed method is architecture-independent and can effectively enhance the performance of various model architectures. Qualitative comparisons based on ResNet-18 are depicted in Figure \ref{fig:pascal_vis}.

\begin{table}[ht]
	\caption{\label{tab: coco-results} Quantitative results on COCO Validation set.}
	\centering
	\resizebox{\linewidth}{!}{%
    \begin{tabular}{>{\raggedright\arraybackslash}m{\dimexpr 0.4\linewidth}|%
                    >{\centering\arraybackslash}m{\dimexpr 0.2\linewidth}%
                    >{\centering\arraybackslash}m{\dimexpr 0.4\linewidth}%
                    }
        \toprule
		Method & Params (M)  & mIoU(\%) \\
		\midrule 
		T: DeepLabV3-ResNet101  & 59.3 & 60.56 \\
		
		\midrule
		
		S: DeepLabV3-ResNet18 &  & 52.08 \\
		S + KD  &   & 54.60 \\
		S + CIRKD  &  16.6 & 55.60\\
		S + DistKD  &   & 55.90 \\
		S + LAD  &   & 56.56 \\
		S + AttnFD (ours)   &   &  \textbf{57.74} \\
		
		\midrule
		
		S: DeepLabV3-MBV2 &  & 47.92 \\
		S + KD  &   & 52.21 \\
		S + CIRKD  & 5.9  & 53.65 \\
		S + DistKD  &   & 53.33 \\
		S + LAD  &   & 55.29 \\
		S + AttnFD (ours)  &    & \textbf{56.95} \\
		
		\midrule
		
		S: PSPNet-ResNet18 &  & 52.68 \\
		S + KD  &  & 54.07 \\
		S + CIRKD  &  12.6 & 56.96 \\
		S + DistKD  &   & 55.06 \\
		S + LAD  &   & 57.50 \\
		S + AttnFD (ours)   &    & \textbf{58.08}   \\
		
		\bottomrule 
	\end{tabular}}
\end{table}

\subsubsection{Cityscapes.}
The proposed method underwent evaluation on the Cityscapes dataset alongside existing methods. Quantitative results presented in Table \ref{tab: cityscapes-results} indicate that AttnFD enhances the performance of ResNet-18, MobileNet, and PSPNet students by 8.95\%, 7.75\%, and 3.14\% respectively. Notably, compared to DistKD and LAD, the second-best methods, AttnFD surpasses them by 1.23\%, 0.96\%, and 0.73\% respectively in terms of mIoU. Additionally, it demonstrates improvements in pixel accuracy, enhancing DistKD and LAD by 2.08\%,  1.1\%, and 0.22\% respectively. The qualitative comparisons utilizing ResNet-18 with DistKD validate the efficacy of the proposed method, as depicted in Figure \ref{fig:city_vis}.

\subsubsection{COCO.}
Similar to the results on the two previous datasets, Table \ref{tab: coco-results} shows the performance of the AttnFD on the validation set of the COCO dataset. The results demonstrate that AttnFD outperforms the SOTA by 1.18\%, 1.66\%, and 0.58\% when using ResNet-18, MobileNet, and PSPNet as the student networks, respectively.

\begin{table}[ht]
	\caption{\label{tab: camvid-results} Quantitative results on CamVid dataset.}
	\centering
        \resizebox{\linewidth}{!}{%
    \begin{tabular}{>{\raggedright\arraybackslash}m{\dimexpr 0.4\linewidth}|%
                    >{\centering\arraybackslash}m{\dimexpr 0.3\linewidth}%
                    >{\centering\arraybackslash}m{\dimexpr 0.3\linewidth}%
                    }
		\toprule
		Method &  \textit{Val} mIoU(\%) & \textit{Test} mIoU(\%) \\
		\midrule 
		T: DeepLabV3-ResNet101  & 76.02  & 65.35 \\
		
		\midrule
		
		S: DeepLabV3-ResNet18 & 71.20 & 62.89 \\
		S + CIRKD  & 76.20 & 67.58 \\
		S + DistKD  & 75.36  & 68.32 \\
		S + LAD  & 76.13 & 66.57 \\
		S + AttnFD (ours)   & \textbf{76.39} & \textbf{68.77}  \\
		
		\midrule
		
		S: PSPNet-ResNet18 & 72.64 & 63.02 \\
		S + CIRKD  & 73.89 & 65.03\\
		S + DistKD  & 75.96  & 65.09 \\
		S + LAD  & 75.84 & 66.13 \\
		S + AttnFD (ours)  & \textbf{76.56} & \textbf{66.74}  \\
		
		\bottomrule 
	\end{tabular}}
\end{table}

\subsubsection{CamVid.}
As shown in Table \ref{tab: camvid-results}, the proposed AttnFD method outperforms existing techniques on both the validation and test sets of the CamVid dataset. The results demonstrate that AttnFD achieves superior performance when using either DeepLab (ResNet-18) or PSPNet (ResNet-18) as the student networks.

\begin{table}[ht]
	\caption{\label{tab: ablation}An ablation analysis conducted on Pascal VOC validation set, examining the influence of distilling refined feature maps across various layers of the network.}
	\centering
        \resizebox{\linewidth}{!}{%
    \begin{tabular}{>{\raggedright\arraybackslash}m{\dimexpr 0.35\linewidth}|%
                    >{\centering\arraybackslash}m{\dimexpr 0.35\linewidth}%
                    >{\centering\arraybackslash}m{\dimexpr 0.35\linewidth}%
                    }
		\toprule
		Method  &  mIoU(\%) &  Accuracy(\%)\\ 
		\midrule 
		T: DeepLabV3-ResNet101  &  77.85 & 86.50\\
		
		\midrule
		S: DeepLabV3-ResNet18 & 67.50 & 76.49 \\
		S + B &  70.25 \textcolor{red}{(+2.75)} & 78.88 \textcolor{red}{(+2.39)}\\
		S + E &   72.31 \textcolor{red}{(+4.81)} & 81.48 \textcolor{red}{(+4.99)}\\
		S + D &    72.47 \textcolor{red}{(+4.97)} & 82.13 \textcolor{red}{(+5.64)}\\
		  S + B + E &  72.58 \textcolor{red}{(+5.08)} & 81.71 \textcolor{red}{(+5.22)}\\
		S + B + D &   72.82 \textcolor{red}{(+5.32)} & 81.87 \textcolor{red}{(+5.38)}\\
		S + E + D &   72.92 \textcolor{red}{(+5.42)} & 82.68 \textcolor{red}{(+6.19)}\\
		S + B + E + D &   73.09 \textcolor{red}{(+5.59)} & 82.95 \textcolor{red}{(+6.46)}\\
		\midrule
		S: DeepLabV3-MBV2 &  63.92 & 73.98\\
		S + B &  66.68 \textcolor{red}{(+2.76)} & 77.01 \textcolor{red}{(+3.03)}\\
		S + E &    68.91 \textcolor{red}{(+4.99)} & 79.60 \textcolor{red}{(+5.62)}\\
		S + D &    69.55 \textcolor{red}{(+5.63)} & 78.50 \textcolor{red}{(+4.52)}\\
		S + B + E & 69.17 \textcolor{red}{(+5.25)} & 79.61 \textcolor{red}{(+5.63)}\\
		S + B + D &   69.46 \textcolor{red}{(+5.54)} & 78.65 \textcolor{red}{(+4.67)}\\
		S + E + D &   69.96 \textcolor{red}{(+6.04)} & 79.73 \textcolor{red}{(+5.75)}\\
		S + B + E + D &  70.38 \textcolor{red}{(+6.46)} & 81.13 \textcolor{red}{(+7.21)}\\
		
		\bottomrule 
	\end{tabular}}
\end{table}

\subsection{Ablation Study}
To further affirm the effectiveness of the proposed method, ablation studies were conducted. Since the method focuses on minimizing the MSE loss between the refined feature maps of the teacher and students, we explored the impact of these feature maps across different layers of the network. Table \ref{tab: ablation} presents the results for three distinct feature maps: "B," which denotes low-level features acquired by the network's backbone, "E," representing features obtained from the last layer of the encoder, and "D," indicating the feature map just before the final convolutional layer of the network. As evidenced by the results, all these features contribute to improved performance, as observed for both ResNet-18 and MobileNet backbones on the Pascal VOC dataset. Notably, the features from the Decoder exhibit more pronounced improvements compared to those from the Encoder, and both surpass the features from the Backbone. This is attributed to the richer, more detailed information contained in the network's final layers. When combined, all three features collectively achieve the best performance.

\begin{figure}[!ht]
	\centerline{\includegraphics[width=\textwidth]{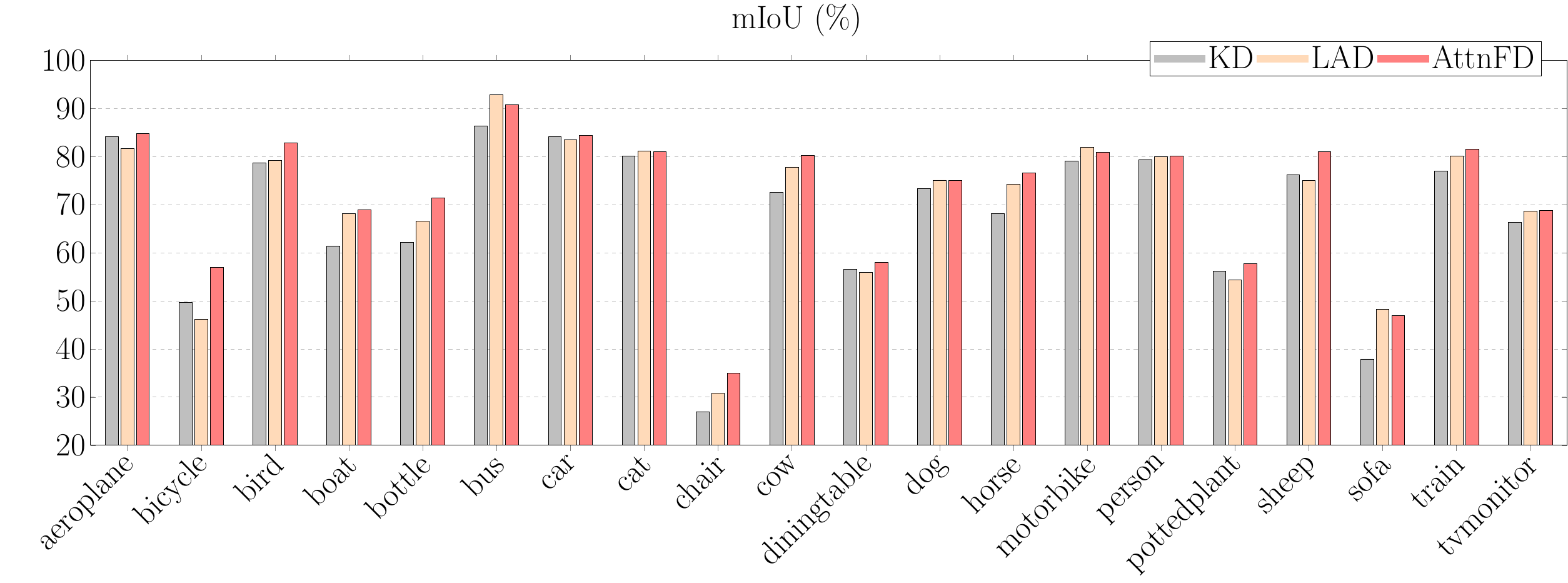}}
	\caption{Visual representation of the performance of proposed method in terms of per-class mIoU using ResNet-18 network on Pascal VOC validation set.}
	\label{fig:pascal_class}
\end{figure}

\begin{figure}[!ht]
	\centerline{\includegraphics[width=\textwidth]{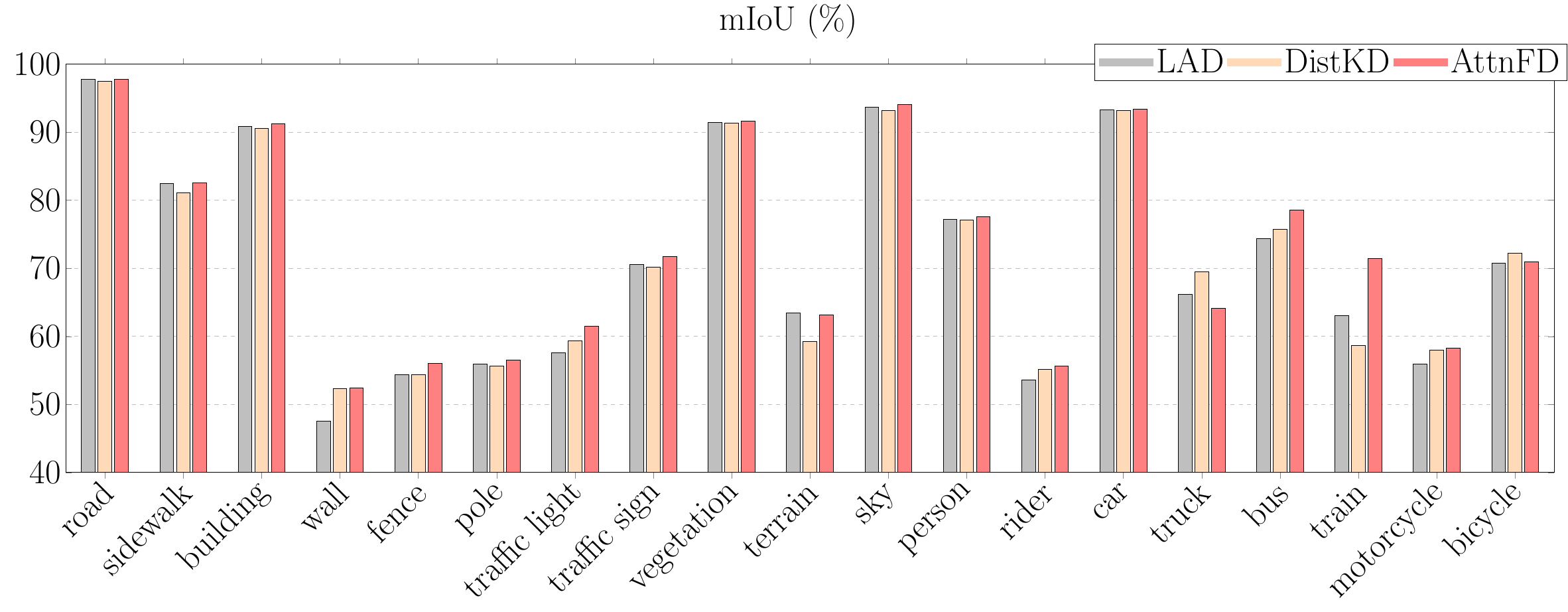}}
	\caption{Comparison of mIoU per class among LAD, DistKD, and AttnFD on Cityscapes validation set, employing a ResNet-18 backbone for the student network.}
	\label{fig:city_class}
\end{figure}

\begin{table}[ht]
	\caption{\label{tab: ablation2} Ablations for different attention modules. Params, Mem, and Time represent the attention mechanism's learnable parameters, memory usage during batched training, and time per image in batched training, respectively. (* denotes that the attention mechanism is applied to one intermediate layer.)}
	\centering
        
        \resizebox{\linewidth}{!}{%
    \begin{tabular}{>{\raggedright\arraybackslash}m{\dimexpr 0.18\linewidth}|%
                    >{\centering\arraybackslash}m{\dimexpr 0.1\linewidth}%
                    >{\centering\arraybackslash}m{\dimexpr 0.15\linewidth}%
                    >{\centering\arraybackslash}m{\dimexpr 0.1\linewidth}%
                    >{\centering\arraybackslash}m{\dimexpr 0.1\linewidth}%
                    >{\centering\arraybackslash}m{\dimexpr 0.6\linewidth}}
		\toprule
		Method & mIoU(\%)  & \#Params & Mem(G) & Time(ms) & Explanation \\
		
		\midrule
		
		S: ResNet-18 & 67.50 & - & 5.48 & 15.71 & w/o attention\\
		S + AT  & 69.60 & - & 13.26 & 58.06 & Channel aggregation w/o learning\\
            S + Lsas  & 70.17 & 52280 & 13.96 & 60.79 & Lightweight sub-attention strategy\\
            S + FcaNet  & 70.69 & 49152 & 14.01 & 61.47 & Multi-spectral frequency channel attention\\
		S + SA  &  $71.72^*$ & $492800^*$  & $12.90^*$ & $32.10^*$ & Pairwise similarity of pixels\\
		S + BAM  & 72.68 & 103235 & 13.72 & 36.88 & Simultaneous channel \& spatial attention\\
		S + EMA  & 72.86 & 3986 & 13.92 & 36.88 & Multi-scale attention by channel grouping \\
		S + CBAM  & 73.09  & 50540 & 14.04 & 62.15 & Channel followed by spatial attention\\
		
		\midrule
		
		S: MobileNet & 63.92 & - & 7.487 & 17.75 & w/o attention\\
		S + AT  & 66.27 & - & 15.35 & 58.74 & Channel aggregation w/o learning\\
            S + Lsas  & 67.09 & 31732 & 16.05 & 61.47 & Multi-spectral frequency channel attention\\
            S + FcaNet  & 67.41 & 29184 & 16.11 & 62.15 & Lightweight sub-attention strategy\\
		S + SA  &  $68.29^*$ & $292880^*$ & $14.97^*$ & $34.83^*$ & Pairwise similarity of pixels\\
		S + BAM  & 69.96 & 61703 & 15.82 & 37.56 & Simultaneous channel \& spatial attention \\
		S + EMA  & 70.05 & 2348 & 16.02 & 37.56 & Multi-scale attention by channel grouping\\
		S + CBAM  & 70.38 & 30368 & 16.11 & 62.84 & Channel followed by spatial attention\\
		
		\bottomrule 
	\end{tabular}}
\end{table}

Figure \ref{fig:pascal_class} illustrates the per-class mIoU comparison between AttnFD and LAD on the Pascal VOC dataset, utilizing ResNet-18 for the student. As depicted, AttnFD performs roughly on par or slightly better than LAD, with notable performance improvements in specific classes, such as bicycle (+10.1), sheep (+5.96), bird (+4.85), and chair (+4.1). As illustrated in the first column of Figure \ref{fig:pull_figure}, the attention maps produced by the proposed method exhibit a stronger focus on the bicycle compared to the raw feature maps, while containing significantly less noise. Furthermore, the visualization of the attention maps and feature maps in the third column demonstrates how the attention mechanism effectively directs more focus towards the main object (the cow) rather than the background, thereby reducing the likelihood of detecting backdoors when segmenting the object.

\begin{figure}[!ht]
	\centerline{\includegraphics[width=\textwidth]{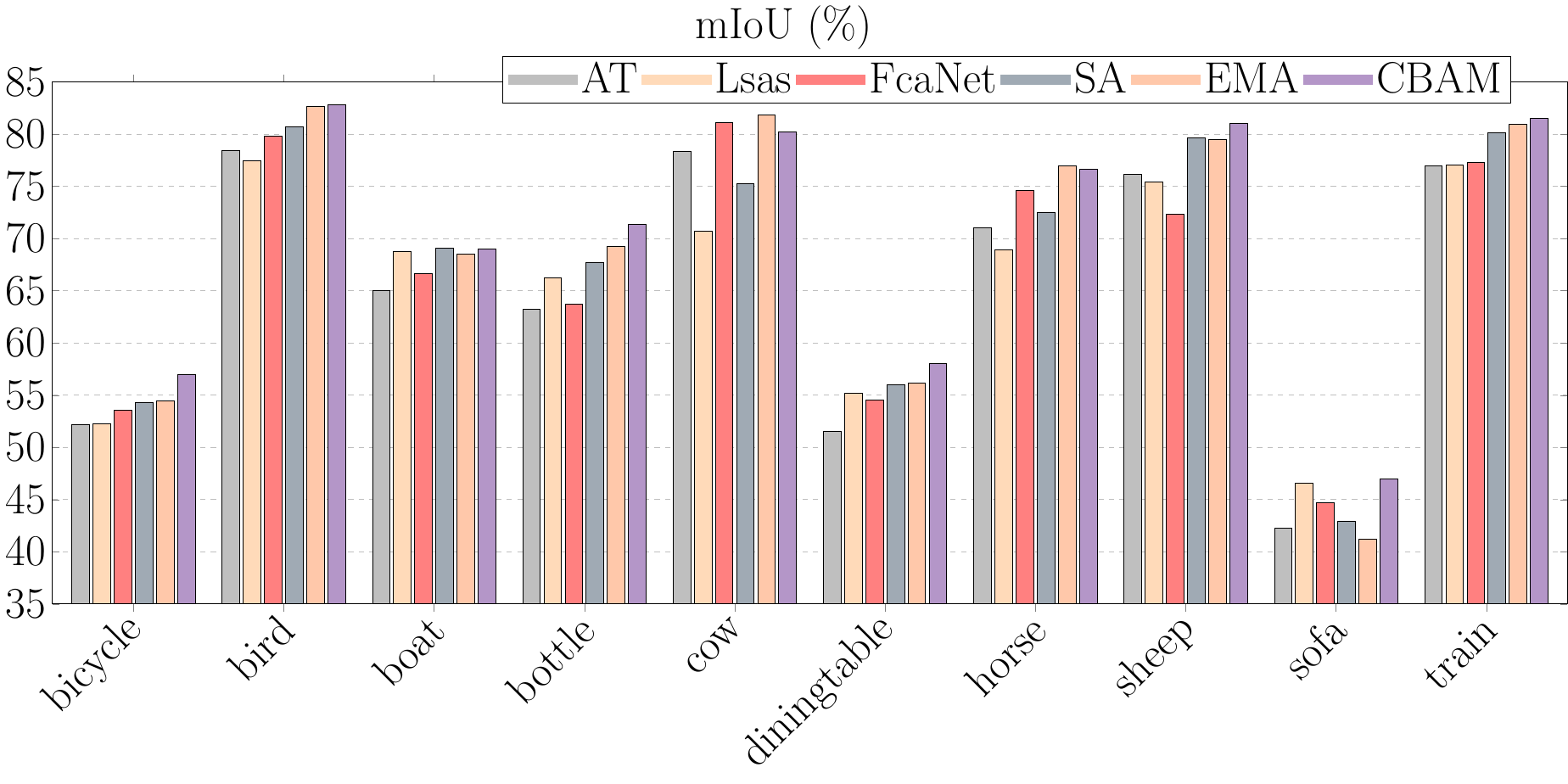}}
	\caption{Comparison of mIoU per class of AttnFD using different attention mechanisms on the Pascal VOC validation set, employing a ResNet-18 backbone for the student network.}
	\label{fig:miou_ablation}
\end{figure}

In a similar vein, Figure \ref{fig:city_class} presents a comparison on the Cityscapes dataset between AttnFD and DistKD (also using ResNet-18 as the student). AttnFD demonstrates significantly superior performance in classes like train (+12.91) and bus (+2.82). The top row of Figure \ref{fig:city_vis} corroborates this, highlighting DistKD's misclassification of the bus as a train, a mistake rectified by AttnFD's improved performance. Moreover, AttnFD exhibits better segmentation of traffic lights and traffic signs compared to DistKD. This improvement can be attributed to the enhanced highlighting of tiny objects like traffic lights in the refined feature maps, which aids the student in mimicking the teacher's attention more effectively, resulting in better segmentation of small objects.

Table \ref{tab: ablation2} presents an ablation study on the different attention modules used for the proposed attention-guided feature distillation on Pascal VOC validation set. This table provides the mIoU improvement, a brief explanation of each method, memory usage in batched training, training time per image in batched training, and the number of learnable parameters for each attention module, namely: Attention Transfer (AT) \cite{zagoruyko2016paying}, Self-Attention (SA) \cite{ambartsoumian2018self}, Bottleneck Attention Module (BAM) \cite{park2018bam}, Frequency Channel Attention Networks (FcaNet) \cite{qin2021fcanet}, Efficient Multi-Scale Attention (EMA) \cite{ouyang2023efficient}, Lightweight sub-attention strategy (Lsas) \cite{zhong2023lsas}, and Convolutional Block Attention Module (CBAM) \cite{woo2018cbam}.

\begin{figure}[!ht]
	\centerline{\includegraphics[width=\textwidth]{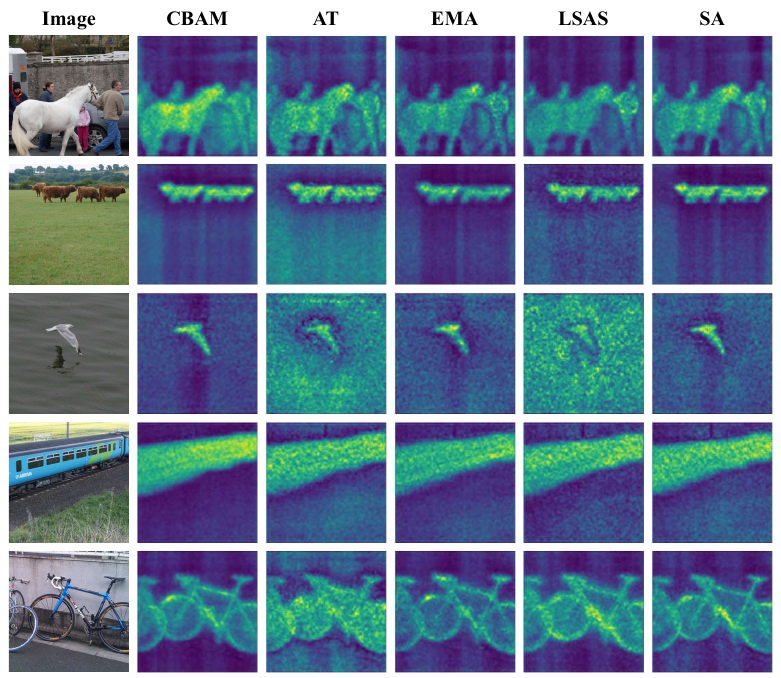}}
	\caption{Visualization of feature maps from different attention mechanisms on the Pascal VOC dataset, employing ResNet-18 as the backbone.}
	\label{fig:att_vis}
\end{figure}

The results show that the SA module contains significantly more learnable parameters, making it more memory-consuming during the training phase. In contrast, the EMA module has the lowest number of parameters (AT module has no learnable parameters at all). However, the CBAM module strikes a balanced trade-off between mIoU improvement and the number of parameters, consistently achieving better results than the other attention modules. CBAM learns to automatically consider both pixel-wise and channel-wise information, which are two important sources of information that have been investigated in the literature for the purpose of knowledge distillation. It should be noted that the results reported for SA apply self-attention on just one layer (denoted by *). Applying self-attention on three intermediate layers would lead to a memory error on our GPU with 24 GB of RAM.

For further validation of the effectiveness of each attention mechanism, Figure \ref{fig:miou_ablation} represents the per-class mIoU of AttnFD using different attention modules for some classes of the Pascal VOC dataset, using ResNet-18 as the backbone of the student network. It can be seen that the performance of each attention mechanism varies by class, and overall, the CBAM and EMA modules perform better than other attention methods. Additionally, Figure \ref{fig:att_vis} visualizes the feature maps of the trained networks with AttnFD, employing different attention methods. It can be observed that CBAM performs better than other methods, as it considers spatial and channel attention, highlighting the most critical regions of the feature and reducing noise. The visualizations are from the last feature map of the student network trained with AttnFD using different attention mechanisms.

\subsection{Discussion}
The proposed method incorporates an additional distillation loss alongside the cross-entropy loss for segmentation. Unlike other existing methods that perform multiple distillation losses, such as the KD loss, it only needs to fine-tune just one hyperparameter. It is worth mentioning that despite our various attempts, combining the proposed method with the KD loss not only did not yield any enhancement in mIoU but it also slightly decreased it  (-0.5\%). This indicates that the proposed loss function effectively distills crucial information from the teacher's refined feature maps to the student. 
Additionally, per-class mIoU results, shown in Figure \ref{fig:pascal_class}, demonstrate that AttnFD consistently surpasses or matches the performance of the KD method across all classes, indicating that incorporating KD loss does not offer additional benefits. In contrast, the LAD method exhibits inferior results compared to the KD method in specific classes (like bicycle and aeroplane). This highlights the necessity of utilizing the KD loss to enhance results for these particular classes in the LAD method.

Furthermore, experimentation with reducing the coefficient of our distillation loss towards the end of training, which has been proven to be helpful in \cite{zhou2020channel, mansourian2025aicsd}, resulted in a minor decrease in performance. This suggests that the CBAM module effectively learns to highlight crucial information for transfer to the student, supported by Figure \ref{fig:pull_figure}, which shows refined feature maps with reduced noise and emphasized regions ready for distillation.

\section{Conclusion}
A novel knowledge distillation method for semantic segmentation was introduced. Unlike existing approaches, which often focus on pairwise information or involve complex distillation losses, the proposed method simplified the process by using raw features and applying channel and spatial attention through the convolutional block attention module to refine feature maps. These refined features highlighted crucial image regions and contained rich information for distillation purposes. Extensive experiments on four known benchmark datasets consistently demonstrated significant performance improvements over models without distillation. Comparison with the state-of-the-art methods further validated the effectiveness of the proposed method.

\section*{CRediT authorship contribution statement}
\textbf{Amir M. Mansourian:} Conceptualization, Methodology, Writing – original draft. \textbf{Arya Jalali:} Conceptualization, Methodology, Writing – original draft. \textbf{Rozhan Ahmadi:} Conceptualization, Methodology, Writing – original draft. \textbf{Shohreh Kasaei:} Supervision, Validation, Writing – reviewing \& editing.

\section*{Declaration of competing interest}
The authors declare that they have no known competing finan- cial interests or personal relationships that could have appeared to  influence the work reported in this paper.

\section*{Data availability}
No datasets were generated. The Pascal VOC 2012, Cityscales, CamVid, and COCO datasets are publicly available

\section*{Funding}
This research did not receive any specific grant from funding agencies in the public, commercial, or not-for-profit sectors.

\bibliographystyle{elsarticle-num}
\bibliography{main}

\end{document}